\documentclass[10pt,twocolumn,letterpaper]{article}

\usepackage[pagenumbers]{cvpr} 

\usepackage[dvipsnames]{xcolor}

\usepackage{amssymb}
\usepackage{pifont}
\newcommand{\xmark}{\ding{55}}
\usepackage[dvipsnames]{xcolor}
\usepackage{threeparttable}
\usepackage[symbol]{footmisc}

\definecolor{cvprblue}{rgb}{0.21,0.49,0.74}
\usepackage[pagebackref,breaklinks,colorlinks,citecolor=cvprblue]{hyperref}
\usepackage{float}

\title{Cubify Anything: Scaling Indoor 3D Object Detection}
\begin{document}
\author{Justin Lazarow \quad David Griffiths$^{\dagger} $\quad Gefen Kohavi \quad Francisco Crespo \quad Afshin Dehghan\\
Apple\quad\quad\quad\quad\quad\\
{\tt\small $\{$jlazarow, d\_griffiths, gkohavi, f\_crespo, adehghan$\}$@apple.com}
}
\twocolumn[{
\maketitle\centering
\captionsetup{type=figure}
\includegraphics[width=0.99\textwidth]{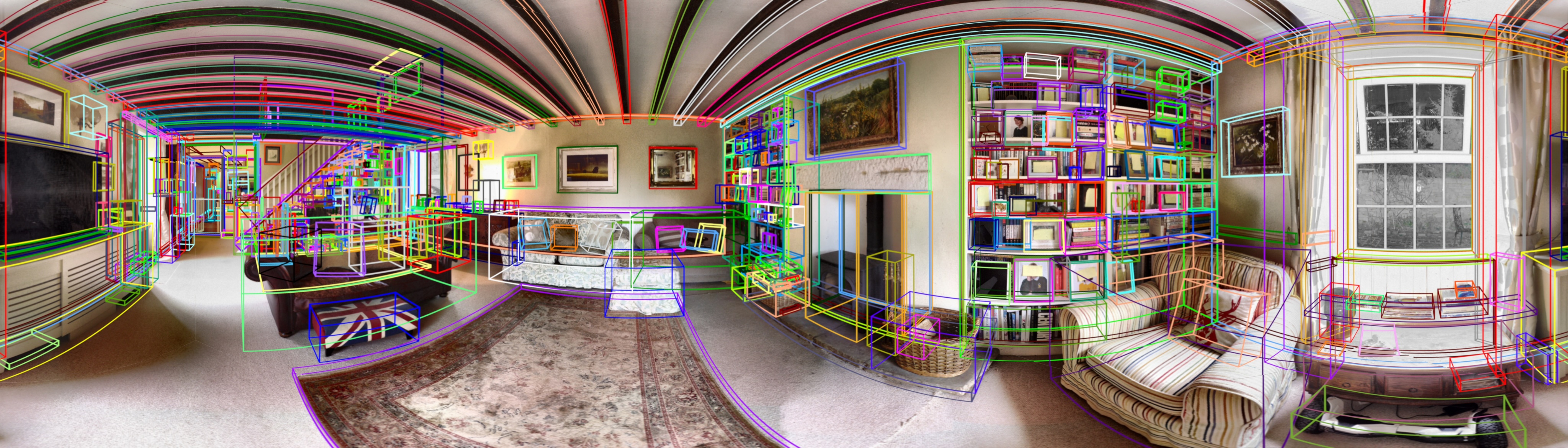}\vspace{-2mm}
\includegraphics[width=0.99\textwidth]{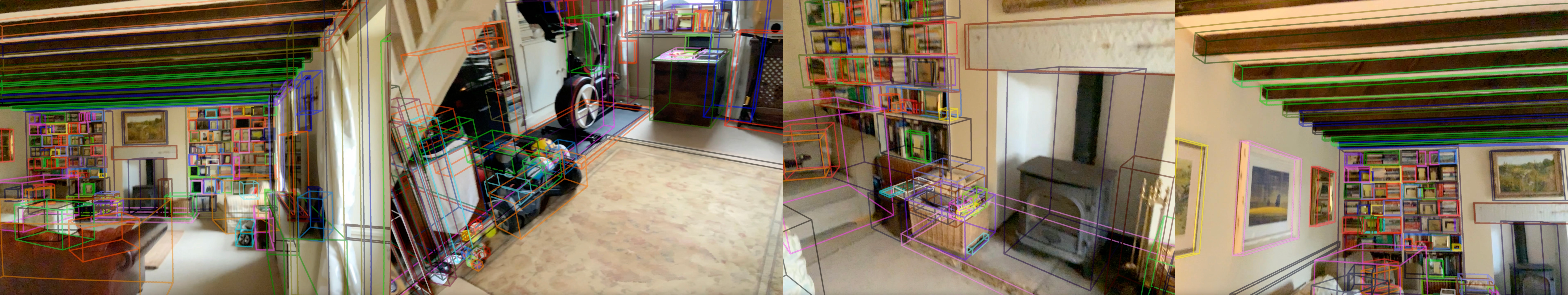}
\captionof{figure}{The Cubify Anything 1M (CA-1M) dataset re-imagines ARKitScenes \cite{baruch2021arkitscenes} by annotating 3D boxes for objects in a near-exhaustive, class-agnostic manner for over 1K of the laser-scanned scenes which have been registered to over 3000 iPad Pro RGB-D captures. We show the richness of these annotations from the perspective of a stationary FARO laser scanner in the panorama (top). The CA-1M dataset subsequently uses the registration to render the annotations in an pixel-accurate manner to \textit{every frame in the capture} as shown in the selected frames of the second row to produce over 15 million frames capturing over 440K objects.}
\label{fig:teaser}\vspace{2mm}
}]
\maketitle
\footnotetext{$\dagger$ Project co-lead}
\begin{abstract}
We consider indoor 3D object detection with respect to a single RGB(-D) frame acquired from a commodity handheld device. We seek to significantly advance the status quo with respect to both data and modeling. First, we establish that existing datasets have significant limitations to scale, accuracy, and diversity of objects. As a result, we introduce the \textbf{Cubify-Anything 1M (CA-1M)} dataset, which exhaustively labels over 400K 3D objects on over 1K highly accurate laser-scanned scenes with near-perfect registration to over 3.5K handheld, egocentric captures. Next, we establish \textbf{Cubify Transformer (CuTR)}, a fully Transformer 3D object detection baseline which rather than operating in 3D on point or voxel-based representations, predicts 3D boxes directly from 2D features derived from RGB(-D) inputs. While this approach lacks any 3D inductive biases, we show that paired with CA-1M, CuTR outperforms point-based methods --- accurately recalling over 62\% of objects in 3D, and is significantly more capable at handling noise and uncertainty present in commodity LiDAR-derived depth maps while also providing promising RGB only performance without architecture changes. Furthermore, by pre-training on CA-1M, CuTR can outperform point-based methods on a more diverse variant of SUN RGB-D --- supporting the notion that while inductive biases in 3D are useful at the smaller sizes of existing datasets, they fail to scale to the data-rich regime of CA-1M. Overall, this dataset and baseline model provide strong evidence that we are moving towards models which can effectively \textbf{Cubify Anything}.
\end{abstract}   
\twocolumn[{
\maketitle\centering
\captionsetup{type=figure}
\includegraphics[width=0.99\textwidth]{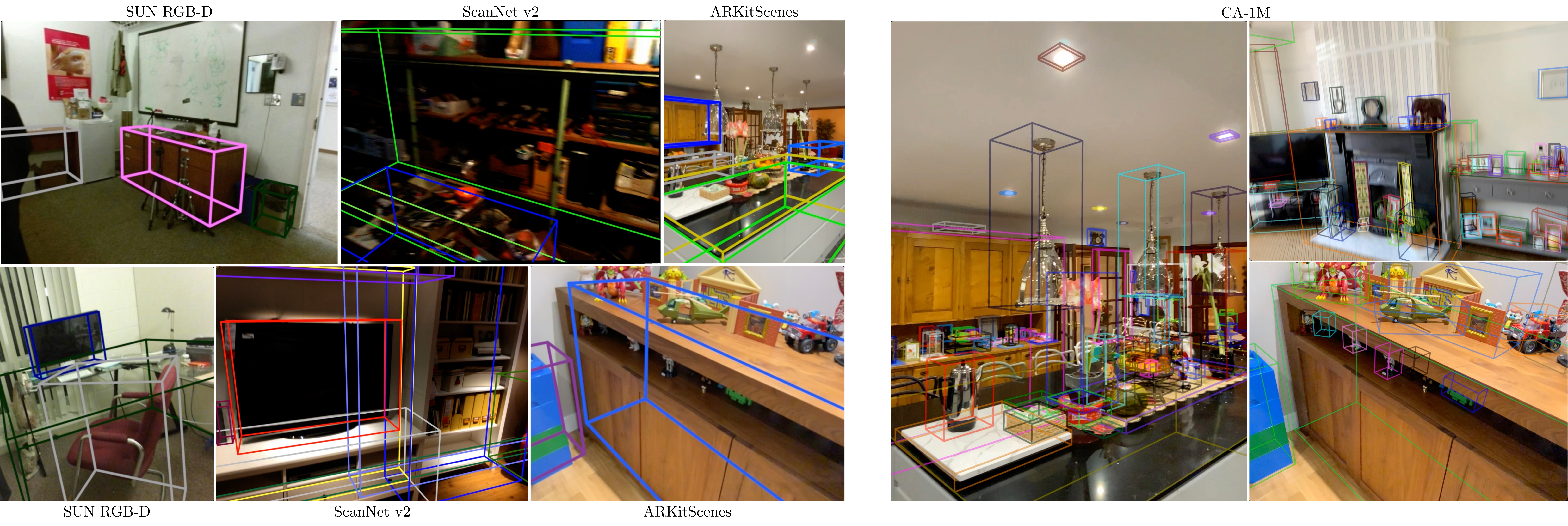}
\captionof{figure}{CA-1M is the first dataset to provide explicit 3D boxes which cover the full richness of objects while being both spatially accurate and pixel-perfect with respect to each frame. Existing datasets like SUN RGB-D, ScanNet v2, ARKitScenes are either small, coarsely labeled, or lack accurate mappings from world to image space. Since ARKitScenes and CA-1M are labeled on the same underlying data, we can show the effect of exhaustive labeling.}
\label{fig:dataset_qual_comp}\vspace{5mm}
}]

\section{Introduction}
\label{sec:intro}

While images are the basis for how we understand the world, indoor 3D object detection has found itself divorced from the source of reality they offer. Commodity sources of depth (ToF, LiDAR) are commonly combined with noisy pose estimates (VIO/SLAM) to realize 3D point clouds or meshes from posed RGB-D images. These 3D representations become not only the source of ground-truth for annotation, but also the inputs to models --- entangling model design with annotation bias. They require complex machinery like sparse operations in 3D to make computation feasible and instill strong inductive biases within models to overcome limited dataset sizes. At the same time, their relatively poor resolution (compared to images) is unable to capture a significant number of objects, especially smaller objects that together enrich our full understanding of a scene. This leads to datasets which are primarily concerned with room-defining objects (chairs, beds, tables) and not the myriad of smaller objects which occupy daily life. Furthermore, given ground-truth annotation on a noisy representation, re-projecting of 3D boxes back to the posed RGB images leads to noticeable misalignment, inconsistency, and jitter. As a result, this has incentivized 3D object detection to be concerned with \textit{scene-level} rather than \textit{image-level} estimates. In this work, we suggest remediation of the aforementioned issues by:

\begin{enumerate}
    \item Building a large-scale, disentangled \textbf{dataset} which accurately annotates every object in both 3D (spatial reality) \textit{and} as projections to every image (pixel-perfect). While the dataset can be used in a scene-level manner, it is primarily presented at the image-level.
    \item Designing a baseline fully Transformer 3D object detection \textbf{model} which has no additional machinery over ordinary 2D object detection architectures.
    \item Re-evaluating whether under this new regime of large, highly accurate, diverse, and disentangled data, the commonly used point/voxel-based models which rely on 3D inductive biases and complexity should still be the \textit{de facto} choice for indoor 3D object detection.
\end{enumerate}

\noindent \textbf{Dataset.} Our dataset, \textbf{C}ubify\textbf{A}nything \textbf{-1M} (CA-1M), extends the underlying LiDAR and handheld RGB-D captures from ARKitScenes \cite{baruch2021arkitscenes} to include 9-DOF 3D boxes for every object in a scene --- expertly labeled on FARO laser scans. While this itself is unique, it is not our ultimate goal. Rather, we use the provided registration between the laser scanners and handheld captures to project the 3D boxes to every frame of each handheld capture. Next, we design a \textit{rendering process} to transform the 3D boxes (which are annotated with respect to the entire scene) to each frame -- reflecting the frustum and occlusion properties from the perspective of that particular frame's viewpoint. This mapping is highly accurate and leads to both 2D and 3D boxes for each frame consistent with what an annotator directly annotating the frame (in 2D or 3D) might produce. The resulting annotations reflect \textit{spatial reality} and at the same time are \textit{pixel-perfect}. Consisting of over 1000 distinct scenes across 3500 handheld captures, this amounts to over 439K unique objects across each video. Notably, each frame of every capture is registered, leading to nearly 13M training frames (i.e., having at least one ground-truth instance) and over 1.8M validation frames.

\noindent \textbf{Model.} Equipped with CA-1M, we seek a model design which can best capture the scale and diversity of objects present. We eschew the design of 3D point or voxel-based detectors which inherently operate in the 3D space and dominate the leaderboards on entangled, smaller scale datasets like SUN RGB-D \cite{song2015sun}. These methods rely on noisily acquired depth maps and can often only compute boxes true to this representation. They require complicated machinery like sparse convolutions or K-nearest neighbor to instill the inductive biases of 3D while making computation feasible. This limits their accessibility --- the mechanisms of voxelization and sparsity are not friendly to many accelerators beyond GPUs. Instead, we seek a design with minimal inductive biases which can effectively scale to the millions of images and pixel-perfect 3D boxes in CA-1M. Therefore, we introduce \textbf{Cu}bify \textbf{Tr}ansformer (CuTR), a fully Transformer-based 3D object detector which operates purely on an RGB image and an optional depth map. CuTR relies on an appropriately pre-trained ViT-based backbone alongside a single-stage/single-scale detector to predict 2D and 3D boxes without ever lifting the inputs into 3D. We show that CuTR, given sufficient training data, can \textit{outperform the best point-based architectures}, not only on CA-1M, but even on small scale, entangled datasets like SUN RGB-D.

\noindent\textbf{Data release.} We are releasing the CA-1M dataset and CuTR models to further aid research within the indoor 3D object detection community. We believe the dataset and approach has extensive applications beyond 3D object detection including: localization and mapping, spatial understanding of MLLMs, and agentic behavior.
\section{Related Work}
\label{sec:rel_work}
We survey and compare CA-1M and CuTR to existing work in both indoor 3D object detection datasets and modeling.
\noindent\textbf{Indoor 3D Datasets.}
Indoor 3D datasets vary in their scale and acquisition methodologies. SUN RGB-D \cite{song2015sun} is one of the earliest and was collected on a variety of available depth sensors (e.g., Kinect V1/V2, RealSense) at the time. Each venue is loosely collected and objects are annotated per-frame with orientation around the gravity axis with classes annotated in a somewhat flexible manner with most work focusing only on a small subset. The primary objects are large and room defining. ScanNet V2 \cite{dai2017scannet} focuses on scene-level reconstruction and was collected using Kinect sensors affixed to an iPad. While it does include instance level segmentation labels, it does not provide explicit oriented 3D box labels. Instead, axis-aligned 3D boxes can be derived by a fitting process and provided axis-aligning transformation. ScanNet++ \cite{yeshwanth2023scannet++} re-imagines ScanNet by annotating similar labels as ScanNet v2 but on high quality FARO laser scans registered to handheld iPhone captures and DSLR images but still lacks explicit 3D box labels as well as axis-aligning transformations. ARKitScenes \cite{baruch2021arkitscenes} uses an iPad and its LiDAR sensor to capture thousands of rooms alongside FARO laser scans. However, while ARKitScenes includes 7-DOF 3D boxes for room-defining objects, these objects are annotated on the handheld LiDAR-derived scene reconstruction rather than the FARO scans.
Overall, each of these existing datasets has caveats on either: size, diversity of objects, annotation source accuracy, or availability of explicit 3D box annotations when compared to CA-1M.

\noindent\textbf{3D Object Detection.}
3D object detection architectures can be categorized by what kind of inputs they require and the manner in which they operate on them. \\
\textit{Point-based methods}: We first consider those which operate on 3D data directly. Early methods were based on point clouds \cite{qi2019deep, qi2020imvotenet} usually operating on them in a permutation-invariant manner. These methods often require new fundamental operations. On the other hand, voxel-based methods \cite{rukhovich2022fcaf3d, rukhovich2023tr3d} operate on a discretization of the 3D space using voxel grids --- serving as an analog to the image-like grid in 2D detection architectures. Convolution can easily be extended to 3D and scaled more efficiently with sparse variants. While it is generally the case that these methods rely on depth (or direct point cloud inputs), some variations \cite{rukhovich2022imvoxelnet} can operate without it by backprojecting entire rays through a dense voxel grid.\\
\textit{Image-based methods:} Monocular approaches \cite{brazil2023omni3d} do not generally use depth as input and instead rely on learned priors to predict 3D boxes directly from images alone. These methods can be multi-stage \cite{brazil2023omni3d} or might directly regress 3D boxes. While our work is also image-based, we primarily focus on the case of a monocular image \textit{alongside} a depth input provided by a commodity LiDAR sensor which allows us to fairly compare to state of the art point-based methods (which always rely on depth).\\
\textit{Scene-level versus image-level:} Additionally, 3D object detection methods are often presented at the \textit{scene-level} (e.g., operating on an entire room/ aggregated representation). We emphasize that we study 3D object detection from the perspective of a single image. Not only is this setting extremely practical, but it removes confounding factors like aggregation of scans and better aligns the task with 2D detection equivalents.\\
\textit{Omni Datasets:} Some recent work \cite{bachmann2022multimae, krishnan2024omninocs} offers ``omni'' solutions to 3D object detection. We note, however, that these approaches seek unified, inter-dataset generalization through dataset aggregation. In our dataset, we seek \textit{intra-dataset} completeness and accuracy in order to produce 3D object detection models which can, for the first time, be supervised to detect \textit{any object}.
\section{CA-1M Dataset}
\label{sec:ca1m}

\begin{table*}[ht!]
\centering
\begin{threeparttable}
\resizebox{.95\textwidth}{!}{
\begin{tabular}{lcccccccc}
\toprule
\textbf{Dataset} & Objects (Categories) & Frames$^\dag$ & Captures & Annotation Source & Spatial Reality & Pixel Perfect & Exhaustive & Boxes \\
\midrule
SUN RGB-D \cite{song2015sun} & 31K (10) & 10K & N/A & IR/ToF (per-frame) & \xmark & \xmark & \xmark & 7-DOF \\
Omni3D SUN RGB-D \cite{brazil2023omni3d} & 52K (38) & 10K & N/A & IR/ToF (per-frame) & \xmark & \xmark & \xmark & 7-DOF \\
ARKitScenes \cite{baruch2021arkitscenes} & 56K (21) & 365K & 5K & ARKit (world) & \xmark & \xmark & \xmark & 7-DOF \\
ScanNet++ \cite{yeshwanth2023scannet++} & 30K (1553) & 264K & 330 & FARO Focus (world) & \checkmark & \checkmark & \xmark & \xmark$^\ddag$ \\
CA-1M & 440K (1) & 15M & 3.5K & FARO Focus (world) & \checkmark & \checkmark & \checkmark & 9-DOF
\end{tabular}}
\caption[caption for table]{Overview of 3D object detection datasets and comparisons to CA-1M. CA-1M introduces an additional order of magnitude of objects through class-agnostic, exhaustive labeling over previous datasets. At the same time, CA-1M is the only dataset that can offer spatial reality (3D accuracy) and pixel perfect (per-frame 2D alignment) alongside explicit 3D box annotations.}
\footnotesize
\begin{tablenotes}
\item[\dag] We only consider frames containing at least one labeled box (i.e., trainable) when considering the size of the dataset
\item[\ddag] While not officially provided, we attempt to derive axis-aligned (i.e., to major walls) 6-DOF boxes for the sake of comparison on ScanNet++
\end{tablenotes}
\label{tab:dataset_comparison}
\end{threeparttable}
\vspace{-4mm}
\end{table*}

The Cubify-Anything 1M (CA-1M) dataset is made up of handheld captures from an iPad Pro and is built on the same underlying raw sensor data from ARKitScenes \cite{baruch2021arkitscenes}. As such, both are made up of over 1000 indoor scenes (i.e., rooms) scanned multiple times and registered alongside multiple FARO laser scans. However, when examining the 3D object detection labels, CA-1M differs in a few key ways:

\begin{enumerate}
\item \textit{Taxonomy:} The 3D boxes in ARKitScenes include only room-defining objects like cabinets, tables, and beds. Conversely, CA-1M aims to exhaustively label \textbf{every and all objects} in a class agnostic manner. While ARKitScenes has 56K objects across 5K videos ($\sim$11 objects per video), CA-1M has over 439K objects across 3.5K videos ($\sim$125 objects per video).
\item \textit{Annotation Source:} In ARKitScenes, the 3D boxes are annotated on noisy, low-resolution meshes acquired purely from the handheld device using ARKit. ARKit provides both estimated depth (assisted by the on-device LiDAR sensor) and estimated pose (using on-device SLAM). Instead of using ARKit meshes, CA-1M is labeled \textit{directly} on the FARO laser scan data. This provides point clouds with survey grade accuracy and resolution which further assists in the ability to label smaller objects. Fundamentally, CA-1M annotations are \textit{disentangled} from any depth or pose estimator and thus reflect an accurate, consistent definition of the 3D world.
\item \textit{Rendering:} Unlike ARKitScenes, ground-truth in CA-1M is produced by \textit{rendering} the world-space annotations to every frame --- producing 2D and 3D boxes which match the frustum and occlusion characteristics of each camera position. Due to the fidelity of the annotation source and registration from \cite{baruch2021arkitscenes}, this rendering is extremely accurate (i.e., pixel-perfect). ARKitScenes contains only annotations on 3D point clouds and attempts at rendering (Figure \ref{fig:dataset_qual_comp}) lead to noisy results which are inherently unfixable.
\end{enumerate}

\noindent Table \ref{tab:dataset_comparison} and Figure \ref{fig:dataset_qual_comp} compare CA-1M to ARKitScenes as well as other existing datasets.

\subsection{Annotation of CA-1M}
\begin{figure}
\centering
\captionsetup{type=figure}
\includegraphics[width=0.45\textwidth]{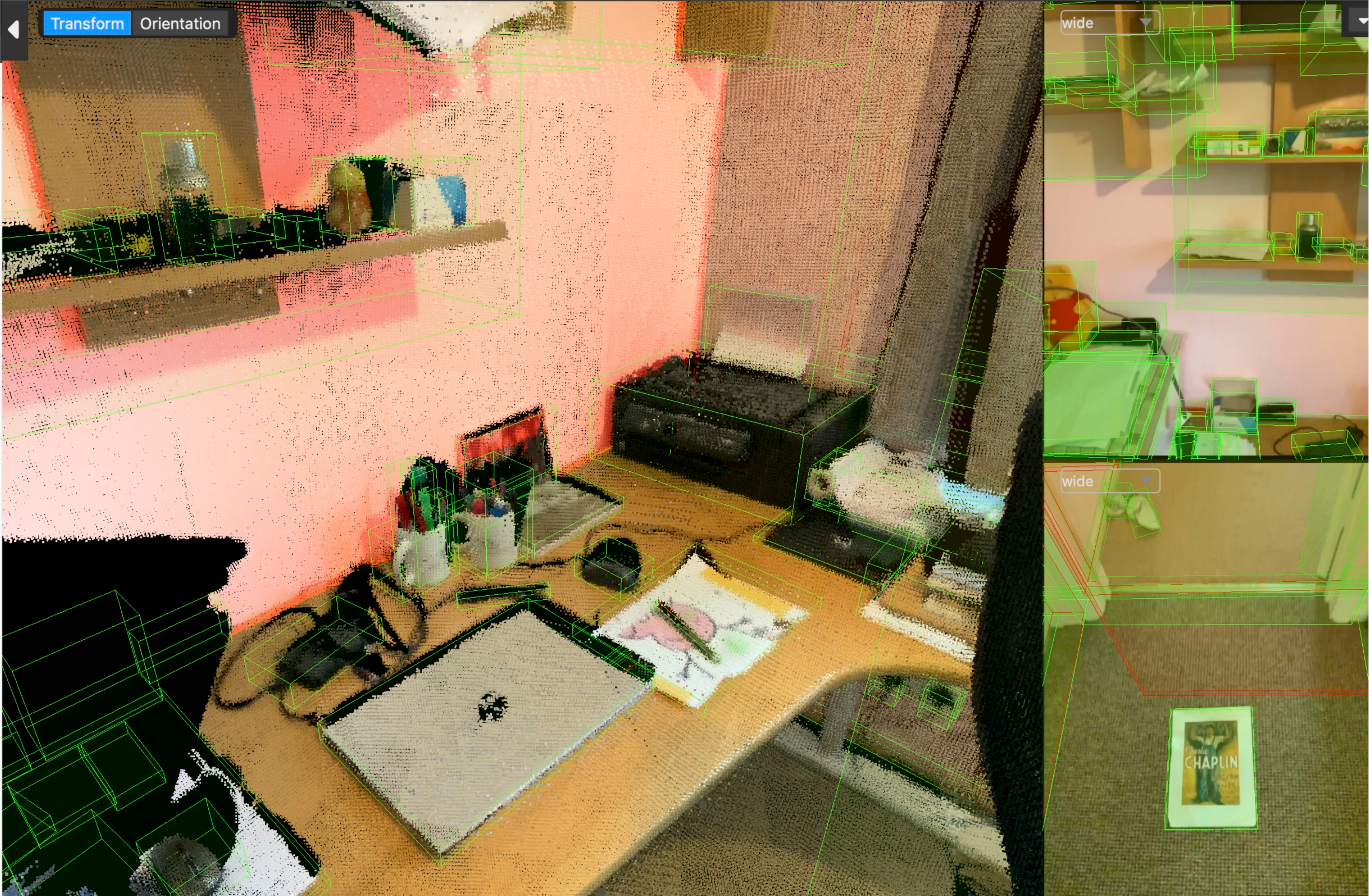}
\vspace{-2mm}
\captionof{figure}{The CA-1M annotation tool targets robustly labeling 3D boxes for \textit{any} object on high-resolution FARO point clouds. Multi-view projection of annotations to supporting images allows for accurate and reliable annotation even when the laser scans include only partial scans of objects, like those on the shelves and to the left of the desk in the accompanying image.
\vspace{-4mm}
\label{fig:annotation_tool}}
\end{figure}

\begin{figure}
\centering
\captionsetup{type=figure}
\includegraphics[width=0.45\textwidth]{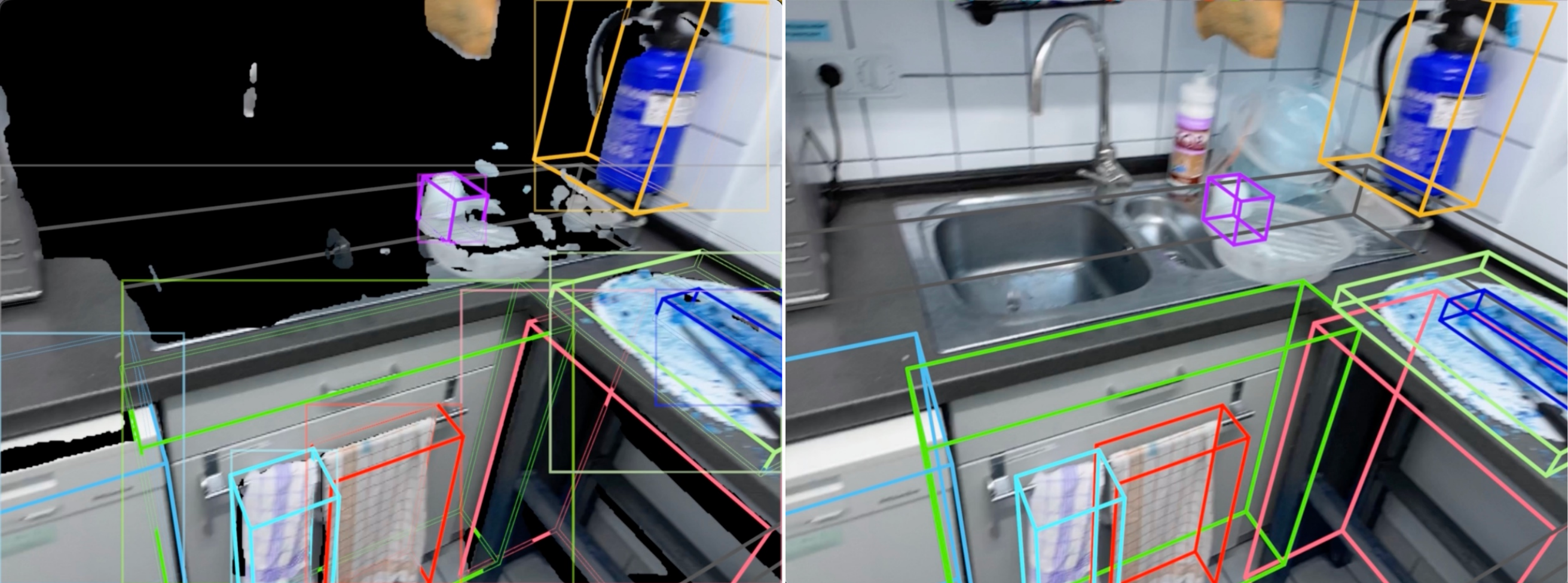}
\vspace{-2mm}
\captionof{figure}{While ScanNet++ is also labeled on FARO scans, it does not explicitly label 3D boxes, instead labeling instance segmentation on the FARO meshes. This presents a mismatch where objects are observed on the corresponding RGB capture but not in the underlying FARO scan. We show renderings of the FARO mesh (left) versus the RGB capture (right) where black regions correspond to missing regions. While CA-1M suffers from the same inherent limitation of stationary laser scanners, its \textit{explicit} annotation of 3D boxes still allows for annotation of objects using multi-view image support, as seen in Figure \ref{fig:annotation_tool}.
\label{fig:spp_missing}}
\vspace{-4mm}
\end{figure}

Annotating objects exhaustively in 3D is nontrivial. Despite having highly accurate laser scans, these scans may be missing certain regions in 3D due to laser scanner placement or sensor limitations with transparent or reflective objects. Previous datasets which also use FARO scans like ScanNet++ ignore such regions (Figure \ref{fig:spp_missing}). Nonetheless, we still expect to label such objects in CA-1M. Therefore, we ensure that our 3D annotation tool (Figure \ref{fig:annotation_tool}) is \textit{aligned} with the underlying handheld video capture by using the registration \cite{baruch2021arkitscenes} between the captures and laser scans. Annotators can not only freely navigate the high resolution 3D point cloud, but they can always overlay a corresponding high resolution RGB frame from the video. Additionally, we attach a series of automatically selected supporting frames to ensure good visual coverage of the object even when the underlying scans may be incomplete. Finally, we include ``model-in-the-loop'' support in our annotation tool which ensures that annotators can efficiently initialize each object by drawing a 2D box on an image using a bootstrapped version of CuTR (Section \ref{sec:cutr}).

\subsection{Rendering CA-1M}
\begin{figure}
\centering
\captionsetup{type=figure}
\includegraphics[width=0.49\textwidth]{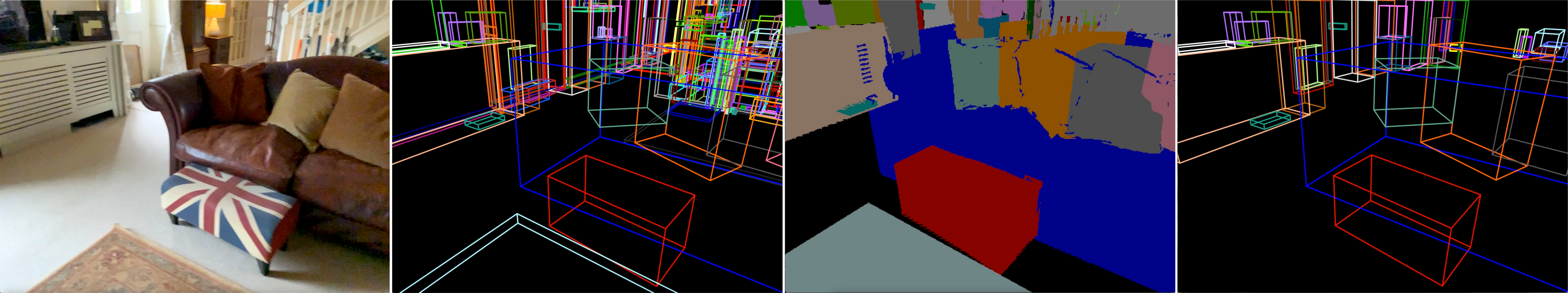}
\captionof{figure}{Per-frame ground-truth is determined by projecting world-space 3D boxes to each frame (left) and uses rendering to determine a coarse ``instance mask'' (middle) which can be used to filter and cut boxes to reflect the frame's visibility and occlusion characteristics (right).}
\label{fig:renderer}
\vspace{-4mm}
\end{figure}

CA-1M is annotated in \textit{world space} like many other 3D datasets (ScanNet, ARKitScenes). However, CA-1M establishes images as the primary representation of the dataset, and therefore we must generate \textit{per-frame} ground-truth. To do so, we design a rendering protocol shown at a high level in Figure \ref{fig:renderer}. Further details are provided in the appendix.

\section{Cubify Transformer}
\label{sec:cutr}
\vspace{-1mm}
\begin{figure*}
\centering
\captionsetup{type=figure}
\includegraphics[width=0.8\textwidth]{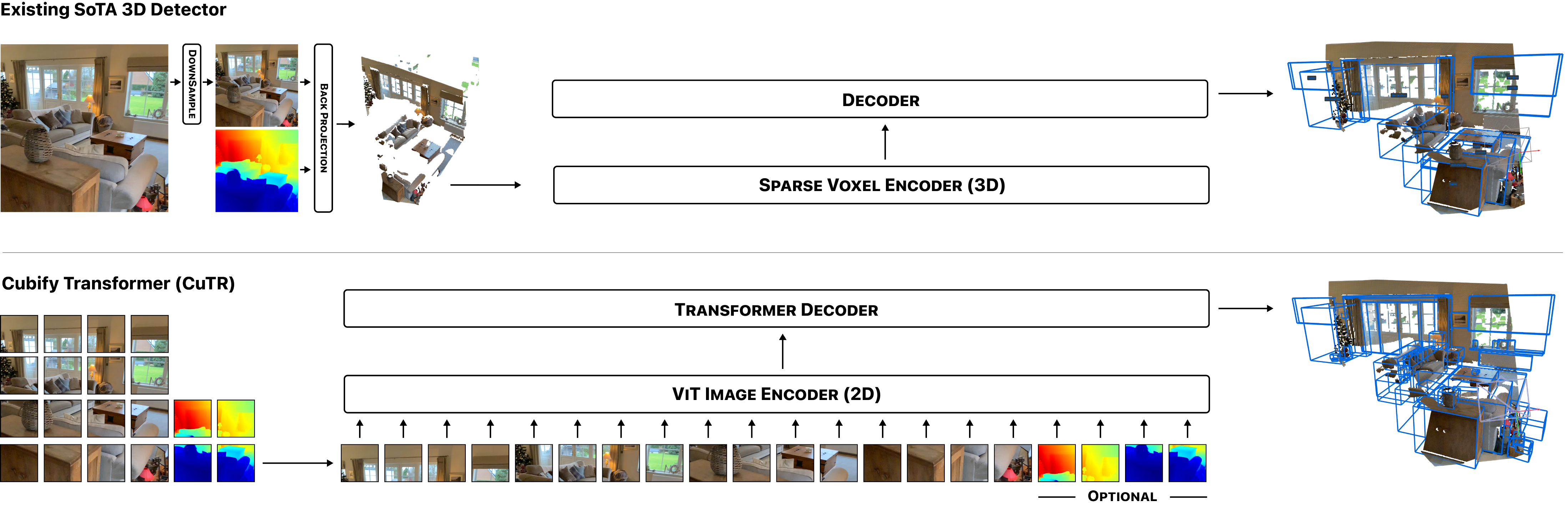}
\vspace{-4mm}
\caption{The architecture of Cubify Transformer (CuTR), compared to an existing SoTA 3D detector architecture. A Vision Transformer \cite{dosovitskiy2020image, bachmann2022multimae} processes both an RGB image and optionally a metric depth map. The resulting detector is built on top of Plain DETR \cite{lin2023detr}, consisting of a single-scale input. Each layer consists of self-attention between decoder queries and global cross-attention modulated by the current 2D box estimate against the feature map. Each query can be decoded to a 2D box, 3D box, and classification score.}
\label{fig:cutr}
\vspace{-4mm}
\end{figure*}

\textbf{Cubify Transformer} (CuTR), extends a single-stage, single-scale transformer-based \cite{vaswani2017attention} 2D detector \cite{lin2023detr} to 3D object detection. We highlight the core choices and justifications which make up the architecture for both RGB-D and RGB variants.

\noindent\textbf{Backbone} For both the RGB only and RGB-D models, we primarily consider CuTR to be best served by a ViT \cite{dosovitskiy2020image} for the effective pre-training they can provide. Each variant is characterized by the underlying ViT and pre-training choices. We use windowed attention to efficiently handle the higher resolution inputs needed for object detection using a base patch size of $16 \times 16$. We do not use relative position embeddings within the windows in order to reduce cost and complexity with negligible loss in overall quality.

\noindent\textbf{RGB-D}
CuTR RGB-D must successfully \textit{fuse} RGB and depth. 
We adopt MultiMAE \cite{bachmann2022multimae}, a ViT-based architecture which is pre-trained with both RGB and affine-invariant depth as a backbone. Both RGB and affine-invariant depth are tokenized as patches before jointly being encoded. Empirically, we find that MultiMAE is capable of operating on asymmetric resolutions (e.g., RGB at $1024 \times 768$ and depth at $256 \times 192$), which we encourage by training on varying ratios of RGB to depth. Practically, this is useful --- RGB generally \textit{is} provided at much higher resolution than a device's depth sensor which helps in minimizing the overall computational cost (e.g., a depth map at $1 / 4$ scale only increases overall tokens by $1 / 16$). Finally, while we encode affine-invariant depth (as expected by MultiMAE), we preserve the scale parameters and subsequently use these to forcibly scale the 3D box predictions (as discussed in \textit{3D Box Predictor} \ref{box_predictor}). Ablations of other backbones are presented in the appendix.

\noindent\textbf{RGB}
CuTR RGB operates purely on RGB without any metric hints. We adopt Depth-Anything \cite{yang2024depth}, a ViT backbone which shows strong affine-invariant monocular depth prediction under massive pre-training as the initialization for our backbone.

\noindent\textbf{3D Box Predictor}
\label{box_predictor}
CuTR requires a strong, accessible 2D object detector and is therefore modeled after the approach of Plain DETR \cite{lin2023detr} ---  a 2D architecture which aims to produce a competitive single-scale DETR-like model through the use of cross-attention which is biased towards a query's predicted 2D box. Notably, this cross-attention is still \textit{global} in nature and thus avoids the complexity and inaccessibility of deformable attention \cite{zhu2020deformable}. CuTR, therefore, adds a direct predictor for 3D boxes. Each query is decoded with an additional MLP which, like Cube R-CNN \cite{brazil2023omni3d}, outputs the projected $xy$ and depth $z$ of the 3D center, the dimensions, and the orientation of the 3D box. Given the known camera intrinsics $K$, the 3D box can be produced by backprojection of the predicted $xy$ and depth $z$ along with the predicted dimensions and orientation. For the RGB-D variant, we used the affine statistics (mean, standard deviation) $\mu$ and $\sigma$ from the metric depth map in order to scale the predicted 3D boxes. Specifically, we re-scale the predicted depth as $z' = \sigma z + \mu$ and predicted dimensions $(l', w', h') = (\sigma l, \sigma w, \sigma h)$.

\noindent\textbf{3D Box Orientation}
We consider CuTR to be \textit{gravity-aware}, and thus we expect the availability of a transformation from the camera to a gravity-aligned coordinate system. This gravity transform is readily available from gyroscope/inertial measurements, and can even be estimated by a neural network \cite{jin2023perspective}. At the same time, while the CA-1M dataset is annotated with 3-DOF orientation (pitch, roll, yaw) \textit{with respect to a gravity aligned} coordinate system, we leave predicting non-zero pitch and roll to future work. Therefore, we supervise CuTR to only predict the yaw-component of the orientation for each 3D box by directly regressing it. For objects in CA-1M which do have pitch and roll, we find the gravity-aligned 7-DOF box which best encloses them during data generation for training and evaluation.

\noindent\textbf{Supervision}
We supervise the corners of the predicted 3D boxes using a Chamfer loss. We find no significant benefit to the disentangled Chamfer loss supervision from Omni3D \cite{brazil2023omni3d}. An ordinary Hungarian matching over the 2D box predictions is used for ground-truth assignment and overall, \textit{CuTR does not rely on NMS}. Avoiding NMS is beneficial --- objects at different depths and sizes might have similar 2D boxes due to perspective effects, and NMS using a 3D IoU would limit the accessibility of the method.

\noindent \textbf{Accessibility} Both variants of CuTR are \textit{accessible}. They are accelerated ``out of the box'' on a multitude of consumer architectures: Metal on Apple Silicon devices, Apple Neural Engine, and ONNX due to reliance only on MLP and vanilla Transformer operations rather than operations like voxelization or sparse convolutions which require custom accelerator kernels.
\section{Experiments}
This section allows us to examine baseline performance of CuTR against existing point-based methods across a series of existing datasets (Section \ref{sec:eval_existing_datasets}): SUN RGB-D (traditional and Omni3D variants), ScanNet++, and CA-1M using a series of models: FCAF3D (point-based), TR3D (point-based). Comparisons of CuTR's RGB only variant are also included against ImVoxelNet and Cube R-CNN. We include some experiments using ScanNet++ in Section \ref{sec:ca1m_as_pretraining}, although we leave more detailed ScanNet++ experiments to the appendix.

Our evaluation on CA-1M (Section \ref{sec:exp_ca1m}) allows us, for the first time, to consider three important questions:

What is the state of ``anything'' indoor 3D object detection? Second, we can ablate, using CA-1M, the effect \textit{entanglement} of previous datasets might have had --- specifically, we ablate methods trained/evaluated on CA-1M with both commodity LiDAR depth \textit{and} high-resolution, ground-truth (FARO-derived) depth (Section \ref{sec:ablate_entanglement}). Finally, we can start to understand the impact of dataset size and quality by re-examining performance on smaller datasets (e.g., SUN RGB-D) under a CA-1M pre-training protocol (Section \ref{sec:ca1m_as_pretraining}).

\noindent \textbf{Metrics.}
The existing literature focuses on class-averaged precision and recall at the traditional 3D IoU thresholds of 0.25 and 0.50. We adopt these same metrics as $AP$ and $AR$ respectively. If classes are not present in the dataset, we use $AP$ and $AR$ in a class-agnostic fashion (i.e., averaged over 1 class). We put a special emphasis on \textbf{a limited number of detections} and all experiments use a limit of 100 detections during evaluation (a standard COCO \cite{lin2014microsoft} practice). Unlike the 2D detection literature, previous evaluations of 3D methods have not generally imposed such limits. While this often has negligible effect on precision, it makes recall evaluation less meaningful and unrealistic when the number of detections is unbounded. 

\begin{table*}[ht!]
    \begin{center}
    \resizebox{.98\textwidth}{!}{
    \begin{tabular}{l|cccc|cccc|cccc|}
        & \multicolumn{4}{c|}{\textbf{Traditional SUN RGB-D}} & \multicolumn{4}{c|}{\textbf{Omni3D SUN RGB-D}} &
        \multicolumn{4}{c|}{\textbf{CA-1M}} \tabularnewline[0.2em]
        \hline
        Method & AP25 & AR25 & AP50 & AR50 & AP25 & AR25 & AP50 & AR50 & AP25 & AR25 & AP50 & AR50 \\
        \hline
        \textit{3D point-based methods} & \\
        \hline
        ImVoxelNet \cite{rukhovich2022imvoxelnet} (RGB only) & 41.0 & 74.9 & 13.5 & 29.0 & 14.4 & 39.0 & 2.5 & 8.8 & 10.1 & 22.8 & 2.3 & 6.3 \\
        \hline
        FCAF \cite{rukhovich2022fcaf3d}  & 63.5 & \textbf{94.2} & 47.0 & 72.5 &  27.1 & 56.5 & \textbf{15.6} & 30.4 & 29.3 & 49.5 & 11.2 & 22.6 \\
        TR3D \cite{rukhovich2023tr3d}  & 66.2 & 93.6 & 49.7 & 72.6 &  27.1 & \textbf{64.2} & 15.2 & 30.9 & 22.0 & 51.9 & 4.4 & 20.0\\
        TR3D + FF \cite{rukhovich2023tr3d}  & \textbf{68.8} & 94.1 & \textbf{51.7} & \textbf{73.7} &  29.1 & 63.3 & 15.5 & \textbf{31.5} & 24.8 & 52.9 & 4.7 & 21.0\\
        \hline
        \textit{2D image-based methods} & \\
        \hline
        Cube R-CNN \cite{brazil2023omni3d} (RGB only) & - & - & - & - &  18.9 & 30.0 & 5.3 & 11.0 & 4.6 & 20.1 & 1.0 & 4.7\\ 
        CuTR (RGB only) & \textit{45.9} & \textit{75.3} & \textit{17.0} & \textit{40.2}  &  \textit{21.5} & \textit{40.4} & \textit{6.9} & \textit{16.9} & \textit{13.5} & \textit{35.4} & \textit{2.4} & \textit{12.9} \\
        \hline
        CuTR (RGB-D) &  59.4 & 87.2 & 34.0 & 56.4 & \textbf{30.3} & 60.2 & 13.6 & 29.0 & \textbf{40.9} & \textbf{62.3} & \textbf{12.7} & \textbf{29.1} \\
        \hline
    \end{tabular}}
    \end{center}
    \vspace{-4mm}
    \caption{We evaluate CuTR against point-based methods on existing 3D datasets. All methods limited to 100 detections. All methods use NMS except CuTR.}
    \label{tab:big_table}
    \vspace{-4mm}
\end{table*}

\subsection{SUN RGB-D}
\label{sec:eval_existing_datasets}

In this section, we focus on establishing a baseline performance of CuTR on existing datasets and against existing models. \textit{No models in this section were trained on CA-1M}.

\noindent\textbf{Traditional SUN RGB-D.}
Table \ref{tab:big_table} shows that \textit{CuTR is a strong approach on the traditional SUN RGB-D split even without consideration of the CA-1M data}. CuTR is competitive with some 3D-based point methods, having only a few point gap in AP25, but remains far behind point-based methods which also use rich 2D features like TR3D+FF. This gap is more significant for AP50 which may show that the inductive biases of existing point-based methods may be well-suited to the larger objects present in the taxonomy. At the same time, we show that when removing depth, the RGB only variant of CuTR is able to significantly outperform the best 3D-based method of ImVoxelNet. This can especially be seen at higher IoU thresholds within AP50 and AR50.

\noindent\textbf{Omni3D SUN RGB-D.}
The traditional SUN RGB-D considers only 10 classes covering large objects. In this section, we consider the Omni3D split, which considers 38 classes and generally includes more small objects. While point-based methods were generally better than CuTR (and 2D-based methods) on the traditional SUN RGB-D split, Table \ref{tab:big_table} shows this may no longer hold for more diverse taxonomies like Omni3D SUN RGB-D. CuTR shows a sizable improvement in AP25 over point-based methods like FCAF and both TR3D with and without 2D feature fusion (FF), and the margin for higher IoU thresholds has narrowed significantly. We consider this as positive signal --- image-based models may perform better as objects become more diverse. CA-1M, therefore, will allow us to take this hypothesis to its limit.

\begin{figure*}
\centering
\captionsetup{type=figure}
\includegraphics[width=0.99\textwidth]{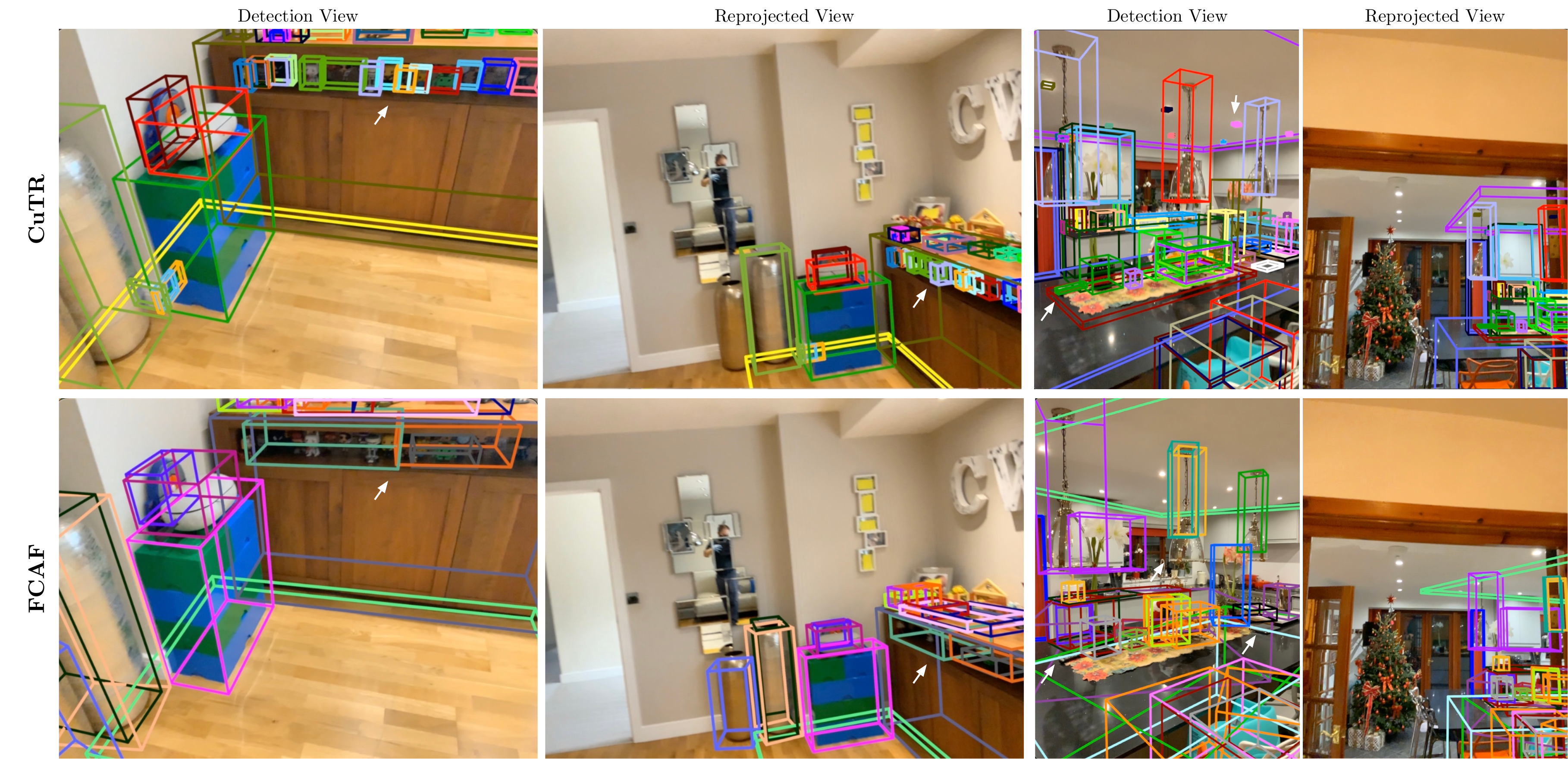}
\captionof{figure}{Visualizations of the qualitative output of CuTR (image-based) and FCAF (point-based) on CA-1M. We show both the frame given to the model (detection view), as well as a frame where these detections are reprojected in order understand depth accuracy. Both outputs are minimally score thresholded (0.075 for FCAF and 0.25 for CuTR). FCAF uses NMS while CuTR has no post-processing. We indicate areas of interest between the two methods.
}
\label{fig:qual_results}
\vspace{-4mm}
\end{figure*}

\subsection{CA-1M}
\label{sec:exp_ca1m}

CA-1M offers a chance to benchmark the current state of ``Cubify Anything'' --- having exhaustively annotated training and validation sets. To ensure fair comparison, no downsampling is performed on point clouds when training point-based models. Each depth map consists of $256 \times 192 \approx 50$K points and voxels are $0.01$m. CuTR is evaluated at an RGB resolution of $1024 \times 768$ and CuTR RGB-D is assisted by a depth map of size $256 \times 192$. The CA-1M results of Table \ref{tab:big_table} tell us that CuTR can recall nearly 62\% of ``all objects'' at the 25\% IoU level (10\% more objects than the best point-based methods). Even without depth, the RGB only variant of CuTR can still recall nearly 35\% of ``all objects'' (almost 13\% more than the best RGB only method). Despite having high recall, CuTR still maintains relatively excellent precision levels. With respect to point-based methods, FCAF appears to be the best, likely owing to its usage of high resolution output features.

\begin{table}[ht]
    \begin{center}
    \resizebox{.35\textwidth}{!}{
    \begin{tabular}{lcccccc}
    \toprule
        Method & AP25 & AR25 & AP50 & AR50  \\
        \midrule
        \textit{3D point-based methods} & \\
        \midrule
        FCAF &  41.3 & 60.9 & \textbf{19.1} & 32.7 \\
        TR3D & 38.2 & 63.6 & 11.0 & 30.4 \\
        TR3D + FF  & 39.2 & 64.5 & 11.4 & 31.2 \\
        \midrule
        \textit{2D based methods} & \\
        \midrule
        CuTR (RGB-D) & \textbf{46.7} & \textbf{67.2} & 15.6 & \textbf{33.1} \\
        \bottomrule
    \end{tabular}}
    \end{center}
    \vspace{-4mm}
    \caption{Ablation on the CA-1M dataset when using high-resolution, ground-truth depth derived from the FARO laser scanner \cite{baruch2021arkitscenes} rather than depth derived from the onboard LiDAR sensor of the iPad Pro.}
    \label{tab:ca1m_gt_depth_eval}
    \vspace{-4mm}
\end{table}

\subsubsection{Explaining (some of) the gap}
\label{sec:ablate_entanglement}
While the results of the previous section are encouraging towards CuTR, they do not provide insight into any specific reason for the gap between image-based and point-based methods on CA-1M. In this section, we argue that this might be explained by point-based methods being less capable of dealing with noisier depth signals. While the previous results study CA-1M in a practical setting --- using LiDAR-derived depth from the iPad Pro, we can also use FARO laser-derived ``ground-truth'' depth. In Table \ref{tab:ca1m_gt_depth_eval}, we observe significant improvements in point-based methods when trained and evaluated on ground-truth depth. While CuTR still continues to outperform the point-based methods in all but AP50, the gap is severely narrowed. This may indicate that point-based models need additional consideration to deal with uncertainty in point clouds --- backprojection and voxelization are relatively ``hard'' operations. Image-based methods, however, can implicitly reason over this uncertainty and use rich image information as a guide. Disentangled datasets like CA-1M allow the empirical study of this phenomena whereas previous datasets merely confound the annotations with the noisily acquired points.

\subsection{CA-1M as Pretraining: \textbf{\textit{SUN RGB-D Revisited}}}
\label{sec:ca1m_as_pretraining}

The SUN RGB-D experiments from Table \ref{tab:big_table} fail to show that CuTR (or image-based methods) can outperform point-based methods on established (albeit entangled) datasets like SUN RGB-D. We use CA-1M as \textit{pre-training} to convincingly show that \textit{SUN RGB-D is simply too small} to be adequately fit without proper pre-training. Therefore, we fine-tune each model, starting from its CA-1M checkpoint, on Omni3D SUN RGB-D in Table \ref{tab:ca1m_omni3d_sunrgbd}. We observe substantial gains across all methods. However, CuTR (both RGB-D and RGB variants) appears to benefit from pretraining significantly more than point-based methods. CuTR shows impressive gains across all metrics, gaining over 13 points in AR25 (and 12 in AR50) which allow it to \textit{significantly outperform point-based methods}. We emphasize that \textit{CA-1M has no semantic labels} and therefore, we believe these gains stem from \textit{more accurate box regression} rather than \textit{better classification}. Furthermore, we can repeat this experiment on CuTR but instead use ARKitScenes and ScanNet++ as pre-training datasets instead of CA-1M. In Table \ref{tab:pretraining_compare}, we see that CA-1M provides significant benefit over both ARKitScenes and ScanNet++, despite CA-1M and ARKitScenes sharing the same underlying data. The scale and accuracy of the oriented 3D boxes in CA-1M provide strong benefits as pre-training, while the lack of inductive biases in CuTR allow for this large scale pre-training to be most effective.
\vspace{-4mm}
\begin{table}[ht]
    \begin{center}
    \resizebox{.49\textwidth}{!}{
    \begin{tabular}{lcccccc}
    \toprule
        Method & AP25 & AR25 & AP50 & AR50  \\
        \midrule
        \textit{3D point-based methods} & \\
        \midrule
        FCAF   &  29.5 \small{(\textcolor{ForestGreen}{+2.4)}} & 65.9 \small{(\textcolor{ForestGreen}{+9.4})} & 18.2 \small{(\textcolor{ForestGreen}{+2.6})} & 39.8 \small{(\textcolor{ForestGreen}{+9.4})}\\
        TR3D   &  28.8 \small{(\textcolor{ForestGreen}{+1.7})} & 63.6 \small{(-0.6)} & 17.0 \small{(\textcolor{ForestGreen}{+1.8})} & 35.4 \small{(\textcolor{ForestGreen}{+4.5})} \\
        TR3D + FF  &  28.8 \small{(-0.3)} & 62.8 \small{(-0.5)} & 16.7 \small{(\textcolor{ForestGreen}{+1.2)}} & 35.3 \small{(\textcolor{ForestGreen}{+3.8)}}\\
        \midrule
        \textit{2D based methods} & \\
        \midrule
        CuTR (RGB only) &  26.0 \small{(\textcolor{ForestGreen}{+4.5})} & 52.4 \small{(\textcolor{ForestGreen}{+12.0})} & 10.1 \small{(\textcolor{ForestGreen}{+3.5})} & 23.9 \small{(\textcolor{ForestGreen}{+7.0})} \\
        \midrule
        CuTR (RGB-D) & \textbf{35.8} \small{(\textcolor{ForestGreen}{+5.5})} & \textbf{73.6} \small{(\textcolor{ForestGreen}{+13.4)}} & \textbf{19.3} \small{(\textcolor{ForestGreen}{+5.7})} & \textbf{40.9} \small{(\textcolor{ForestGreen}{+11.9})} \\
        \bottomrule
    \end{tabular}}
    \end{center}
    \vspace{-4mm}
    \caption{Results of an initialization from a pre-trained CA-1M model on the Omni3D SUN RGB-D split. Whereas point-based models outperformed CuTR in higher threshold metrics in Table \ref{tab:big_table}, CuTR surpasses all methods given sufficient pretraining using CA-1M.}
    \label{tab:ca1m_omni3d_sunrgbd}
    \vspace{-4mm}
\end{table}
\begin{table}[ht]
    \begin{center}
    \resizebox{.35\textwidth}{!}{
    \begin{tabular}{lcccccc}
    \toprule
        Pre-training & AP25 & AR25 & AP50 & AR50  \\
        \midrule
        ARKitScenes & 28.3 & 63.4 & 14.4 & 32.4 \\
        ScanNet++ & 31.4 & 69.9 & 15.9 & 36.3 \\
        CA-1M & 35.8 & 73.6 & 19.3 & 40.9 \\
        \bottomrule
    \end{tabular}}
    \end{center}
    \vspace{-4mm}
    \caption{CA-1M provides the most benefit as a pre-training dataset compared to ARKitScenes and ScanNet++ when fine-tuning CuTR on Omni3D SUN RGB-D.}
    \label{tab:pretraining_compare}
    \vspace{-4mm}
\end{table}
\subsection{Qualitative Results}
We show qualitative results comparing CuTR and FCAF on CA-1M in Figure \ref{fig:qual_results}. Each model is run on the frame (``detection view'') and the results are additionally re-projected to a nearby frame (``reprojected view'') to allow for consideration of depth accuracy. The results demonstrate the advantages of CuTR over FCAF. CuTR accurately captures significantly more objects (especially smaller ones), handles objects with transparency well, and produces detections with an overall better alignment to the image than FCAF. FCAF struggles to produce a holistically satisfying output --- missing (or grouping) many small and thin objects while still producing significant false positives and struggling to correctly predict box rotations.

\section{Conclusion}
We presented both a new dataset (CA-1M) and new model (CuTR) which challenge the idea that point-based 3D architectures are better suited for indoor 3D object detection. While point-based methods might appear better on smaller, less diverse datasets like SUN RGB-D, this becomes less apparent on more diverse datasets like Omni3D SUN RGB-D. Furthermore, in the extreme case of CA-1M, which exhaustively annotates objects and is annotated on highly accurate laser scans rather than point-clouds derived from noisy depth and pose, image-based models like CuTR show significant improvements in both recall and precision. We believe this supports a future with image-based methods rather than point-based methods --- with CA-1M serving as a strong dataset to facilitate such research.

{
    \small
    \bibliographystyle{ieeenat_fullname}
    \bibliography{main}
}

\clearpage
\setcounter{page}{1}
\maketitlesupplementary

\section{CA-1M}
\subsection{Additional Dataset Samples}

We include additional panoramic projections of sample annotations in Figure \ref{fig:more_samples}.

\subsection{Rendering CA-1M}
\label{sec:render_process}

We provide more details on the process to generate CA-1M per-frame ground truth. Given a video capture consisting of N frames, per-frame intrinsics $K_i$, possible distortion parameters $d_i$, and per-frame pose ${RT}_i$ registered to the same coordinate system as the laser scans, we define a high-level rendering pipeline:

\begin{enumerate}
    \item Transform 3D boxes from world to camera space using ${RT}_i$.
    \item Apply frustum culling, removing boxes irrelevant to the current camera pose and frustum
    \item \label{{item:render}} Render each box independently at a suitable resolution (e.g., 320x240) using $K_i$ and $d_i$.
    \item Cut each 3D box to the \textbf{frustum} by backprojecting the independently rendered masks along rays terminating at the far position of the frustum.
    \item \label{item:coarse_viz} Determine coarse visibility of each box by considering the the subset of each rendered mask whose depth is closest to the camera.
    \item Cut each 3D box to the \textbf{scene} by backprojecting the visibility masks along rays terminating at the rendered \textit{depth} to solve for visibility and occlusion.
    \item Remove 3D boxes which have had too significant of a cut relative to the original box.
\end{enumerate}

\noindent This process was shown in the main paper as Figure \ref{fig:renderer}. This pipeline is implemented in PyTorch3D \cite{ravi2020pytorch3d} and parallelized across multiple GPUs and nodes such that ground-truth can feasibly be generated for every frame over thousands of captures.

\subsubsection{Importance of Handling Distortion}

\begin{figure}
\centering
\captionsetup{type=figure}
\includegraphics[width=0.48\textwidth]{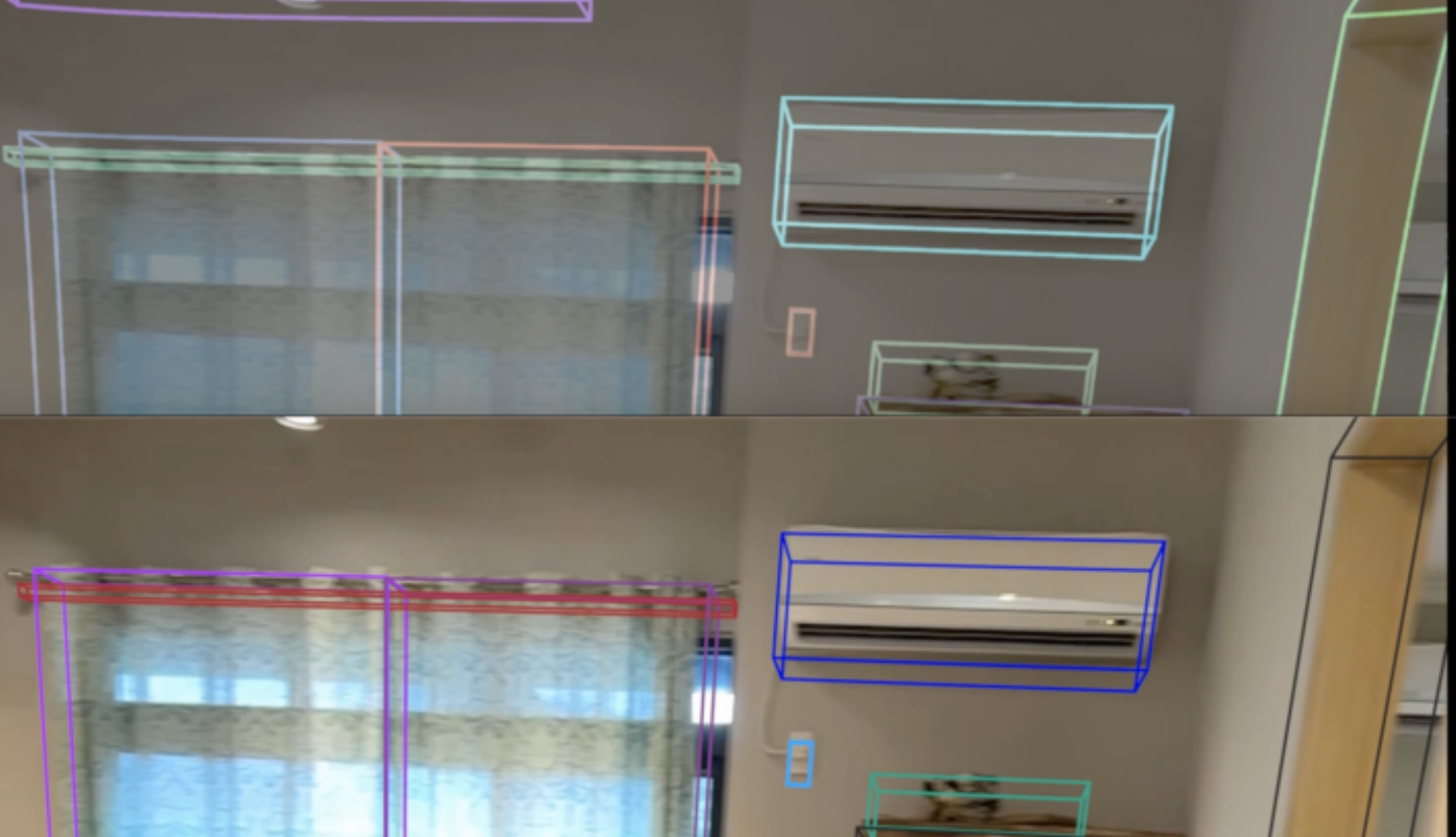}\
\captionof{figure}{Rendering 3D boxes with distortion is important on CA-1M in order to maintain pixel accuracy and alignment for small, distant objects. The top image is rendered with distortion whereas the bottom is rendered purely using a Pinhole model.
\label{fig:distortion_diff}}
\end{figure}

During rendering (of boxes), we must handle distortion (even in the main camera) within the handheld captures in order to maintain pixel alignment, as shown in Figure \ref{fig:distortion_diff}. While this is easier to implement when dealing with points or high resolution meshes, this is non-trivial to implement for 3D boxes.

\section{Cubify Transformer}

\subsection{Additional Training Details}

All image-based methods (CuTR, Cube R-CNN) are trained in a framework based on Detectron2. CuTR uses settings similar to ViTDet (AdamW optimizer, layer decay rates) while we use the same settings from Omni3D \cite{brazil2023omni3d} for traning Cube R-CNN models. All point-based methods are trained in MMDetection3D using the default settings from each respective work.

Image-based models are trained across $8 \times 4$ V100 nodes using a total batch size of 64 (2 images per GPU). All data is processed into WebDataset format and streamed to each node --- keyframing (with some randomness) is applied on the fly. Images are randomly augmented using the ViTDet style augmentations: a random factor between 0.25 and 1.25 is selected and multiplied by a base resolution of $1024 \times 1024$ to get the target image size. A random crop is taken (or the full image if the resolution is smaller). The depth map is augmented accordingly to preserve the same ratio of image to depth. We also select to randomly augment the image to depth ratio (1, 2 or 4). Random horizontal flipping is applied.

\section{ScanNet++}

\subsection{ScanNet++ versus CA-1M}
ScanNet++ adopts the same FARO scanner and handheld capture approach as ARKitScenes/CA-1M and thus can, in theory, offer similar accuracy in 3D annotations to CA-1M. However, we note significant differences compared to CA-1M. First, ScanNet++ is \textit{not a detection dataset}, only offering semantic instance segmentation of a large but not exhaustive taxonomy on the FARO meshes. While we can (and do) attempt to derive 6-DOF axis-aligned boxes from these, annotating with explicit 3D boxes ensures that \textit{every object} can labeled, as done in CA-1M. This is because even a high quality scanner like FARO will still be unable to acquire points for many objects in a scene (field-of-view, material, etc). Points will inevitably be missing and so by only labeling acquired points as ScanNet++ does, a significant number of objects will still go unlabeled (see Figure \ref{fig:spp_missing}). Second, ScanNet++ is an order of magnitude smaller than CA-1M in terms of scans and captures --- having 230 captures versus 3,500 in CA-1M. Finally, the registration process in ScanNet++ is less complete --- \textit{only able to directly register a subset} of frames (264K out of 3.2M frames, i.e., 8\%) to the FARO space. As mentioned in ScanNet++ \cite{yeshwanth2023scannet++}, frames are considered not registerable when the rendering of depth from the FARO scans does not sufficiently agree with depth acquired from the handheld device (i.e., the LiDAR-equipped iPhone). Not only does this lead to less available frames, it \textit{biases} the dataset towards frames where the handheld's acquired depth is ``accurate'' --- partially entangling the dataset to the sensors of the handheld device. On the other hand, the registration process of ARKitScenes/CA-1M registers nearly every frame across all captures.

\subsection{Experiments on ScanNet++}

While we don't attempt to draw any conclusions dissimilar from the experiments on CA-1M, we, for completeness, include results of models trained on ScanNet++ data. Since ScanNet++ does not offer 3D box annotations, we must derive them. We use the 3D instance segmentation labels and use the normal of the first ``wall'' object in each scan to define the basic rotation of the scene and thus produce 6-DoF (axis-aligned) 3D boxes. Notably, this can be error prone if this wall happens to not adhere to some Manhattan assumption, which happens rarely. To generate per-frame ground-truth, we augment the same protocol defined in Section \ref{sec:render_process}, but instead of using the rendered 3D boxes to generate a coarse instance mask (like in Figure \ref{fig:renderer}), we render the modal instance segmentation from the underlying mesh annotations. The rest of the process (filtering, cutting, etc) proceeds the same as CA-1M. While ScanNet++ has a significant number of class labels, we do not consider them in experiments and treat this as a class-agnostic dataset to avoid complications that may arise due to the vocabulary size. Experimental results are presented in Table \ref{tab:scannetpp}.

\label{sec:exp_scannetpp}
\begin{table}[ht]
    \begin{center}
    \resizebox{.5\textwidth}{!}{
    \begin{tabular}{lcccccc}
    \toprule
        Method & AP25 & AR25 & AP50 & AR50  \\
        \midrule
        \textit{3D point-based methods} & \\
        \midrule
        ImVoxelNet (RGB only)  & 21.3 & 40.7 & 4.4 & 11.7 \\
        \midrule
        FCAF &  41.1 & 65.3 & 16.4 & 33.2 \\
        TR3D & 38.1 & 68.3 & 12.8 & 30.9 \\
        TR3D + FF  & 40.3 & 66.6 & 14.1 & 31.4 \\
        \midrule
        \textit{2D based methods} & \\
        \midrule
        Cube R-CNN (RGB only) & 15.5 & 33.2 & 3.5 & 10.6 \\ 
        CuTR (RGB only) &  25.8 & 49.0 & 7.0 & 20.2 \\
        \midrule
        CuTR (RGB-D) & \textbf{48.7} & \textbf{71.5} & \textbf{18.5} & \textbf{36.6} \\
        \bottomrule
    \end{tabular}}
    \end{center}
    \caption{Results on the ScanNet++ dataset when using ARKit depth and axis-aligned (i.e., to walls) ground-truth boxes derived from provided instance segmentation. We evaluate the class-agnostic precision and recall of each method.}
    \label{tab:scannetpp}
\end{table}

Generally, we observe similar trends as we see on CA-1M: CuTR (both RGB and RGB-D variants) outperforms existing methods over the range of evaluation thresholds.

\section{ARKitScenes}

We present no specific results for ARKitScenes, however, we do use a CuTR model pre-trained on per-frame ARKitScenes data in order to conduct the experiments in \ref{tab:pretraining_compare}. For uniformity, this per-frame ground-truth is generated using the same protocol as CA-1M, with only the underlying pose and 3D box annotations coming from the original ARKitScenes \cite{baruch2021arkitscenes} dataset.

\section{Further Ablations}

We provide additional ablations which attempt to answer not only specific questions about CuTR and CA-1M but also to further ablate the shortcomings of point-based methods.

\noindent\textbf{Model Size}
While all experiments of CuTR are done using a ViT-B backbone, we additionally ablate how CuTR responds to smaller backbones: a ViT-S (6 heads, 384 dimension embedding) and a ViT-T (3 heads, 192 dimension embedding). We observe that CuTR scales to these smaller backbones gracefully, still retaining performance competitive with point-based methods even at the ViT-T size.

\begin{table}[ht]
    \begin{center}
    \resizebox{.4\textwidth}{!}{
    \begin{tabular}{lcccccc}
    \toprule
        ViT & AP25  & AR25 & AP50 & AR50  \\
        \midrule
        Tiny & 28.8 & 54.3 & 7.3 & 22.0 \\
        Small & 34.8 & 58.3 & 9.6 & 25.4 \\
        Base & 40.9 & 62.3 & 12.7 & 29.1 \\
        \bottomrule
    \end{tabular}}
    \end{center}
    \caption{CuTR can still retain competitive performance on CA-1M while using significantly smaller model sizes like ViT-S and ViT-T.}
    \label{tab:cutr_ablate_model_size}
\end{table}

\noindent\textbf{Depth (Modality) Fusion}
CuTR adopts a MultiMAE-like backbone: a separate depth patch embedding and joint encoding of both RGB and depth tokens within a Transformer encoder. This can be initialized by a pre-trained MultiMAE (i.e., on pseudo-labeled ImageNet) or we could use a strong RGB only backbone like DINOv2 or Depth-Anything and learn the depth fusion (from scratch) during training. In Table \ref{tab:cutr_ablate_backbone_init}, we show the benefit of initialization from MultiMAE pre-training over DINOv2 \cite{oquab2023dinov2} or Depth-Anything \cite{yang2024depth}, despite these being considered much stronger backbones than MultiMAE. We compare both the 2D and 3D box accuracy which helps explain whether performance differences come from 2D box quality or from 3D.

\begin{table}[ht]
    \begin{center}
    \resizebox{.5\textwidth}{!}{
    \begin{tabular}{lcccccc}
    \toprule
        Backbone & AR50 (2D) & AR75 (2D) & AR25 (3D) & AR50 (3D)\\
        \midrule
        MultiMAE \cite{bachmann2022multimae} & 83.6 & 55.1 & \textbf{62.4} & \textbf{29.0} \\
        DINOv2 \cite{oquab2023dinov2} & \textbf{83.9} & \textbf{55.2} & 60.5 & 27.2 \\
        Depth-Anything \cite{yang2024depth} & 83.8 & 55.0 & 59.8 & 26.7 \\
        \bottomrule
    \end{tabular}}
    \end{center}
    \caption{CuTR initialized with a pre-trained MultiMAE backbone shows strong performance on CA-1M against pre-trained DINOv2 and Depth-Anything backbones which must learn the depth modality fusion during training. While all share similar 2D box performance, MultiMAE shows significant improvements with respect to 3D accuracy. This may show the importance of depth fusion at \textit{pre-training} time for 3D box prediction.}
    \label{tab:cutr_ablate_backbone_init}
\end{table}

\noindent\textbf{``Pixel for point''}
All point-based methods are inherently limited by the depth provided by the underlying sensor, i.e., they can only backproject at a resolution limited by the depth ($256 \times 192$ in CA-1M). In practice, depth is usually significantly lower resolution than what the RGB sensor is capable of delivering ($1024 \times 768$ in CA-1M). Therefore, an advantage of CuTR could be seen as efficiently fusing lower resolution depth data with higher resolution RGB data. In Table \ref{tab:cutr_pixel_for_point}, we purposefully decrease the RGB resolution given to CuTR to understand how ``pixel for point'' it compares to point-based methods. Even at the exact ``pixel for point'' setting of RGB and depth at $256 \times 192$, CuTR can match the performance of point-based methods. Restoring this resolution even to $512 \times 384$ shows minimal losses over the full $1024 \times 768$ evaluation setting of CuTR.

\begin{table}[ht]
    \begin{center}
    \resizebox{.5\textwidth}{!}{
    \begin{tabular}{lcccc}
    \toprule
        RGB Res. & AR50 (2D) & AR75 (2D) & AR25 (3D) & AR50 (3D)  \\
        \midrule
        $1024 \times 768$ &  83.6 & 55.2 & 62.3 & 29.1\\
        $512 \times 384$ & 81.7 & 52.2 & 59.3 & 26.0 \\
        $256 \times 192$ & 81.1 & 51.9 & 50.7 & 19.5 \\
        \midrule
        \textit{3D point-based methods} & \\
        \midrule
        FCAF & N/A & N/A & 49.5 & 22.6 \\
        TR3D & N/A & N/A & 51.9 & 20.0 \\
        TR3D + FF & N/A & N/A & 52.9 & 21.0 \\
        \bottomrule
    \end{tabular}}
    \end{center}
    \caption{CuTR is evaluated on CA-1M at a variety of RGB resolutions with a fixed depth resolution of $256 \times 192$. We reproduce the point-based method results here for convenience. Both 2D and 3D recall suffers at lower RGB resolutions, however, CuTR remains competitive with point-based methods.}
    \label{tab:cutr_pixel_for_point}
\end{table}

\noindent We further understand how the resolution affects performance over \text{distance} by decomposing the 3D evaluation of Table \ref{tab:cutr_pixel_for_point} by \textbf{box distance}. We group all boxes into \textit{near} (0-2 meters), \textit{medium} (2-4 meters), and \textit{far} (4-5 meters) buckets. 

\begin{table}[ht]
    \begin{center}
    \resizebox{.4\textwidth}{!}{
    \begin{tabular}{lccc}
    \toprule
        RGB Res. & AR25n & AR25m & AR25f \\
        \midrule
        $1024 \times 768$ &  68.4 & 58.8 & 40.0\\
        $512 \times 384$ & 66.4 & 54.5 & 35.4 \\
        $256 \times 192$ & 59.1 & 44.5 & 24.8 \\
        \bottomrule
    \end{tabular}}
    \end{center}
    \caption{We view the results of Table \ref{tab:cutr_pixel_for_point} with the 3D evaluation broken down by \textbf{n}ear, \textbf{m}edium, \textbf{f}ar box distances.}
    \label{tab:cutr_pixel_for_point_distance}
\end{table}

We observe that the most significant drops in performance as RGB resolution decreases, comes from the further objects (medium and far distances), which may stem from the possibility that both depth uncertainty increases and size in image decreases as objects are further away. Higher RGB resolutions would better allow the network to learn to overcome these shortcomings.

\twocolumn[{
\maketitle\centering
\captionsetup{type=figure}
\includegraphics[width=\textwidth]{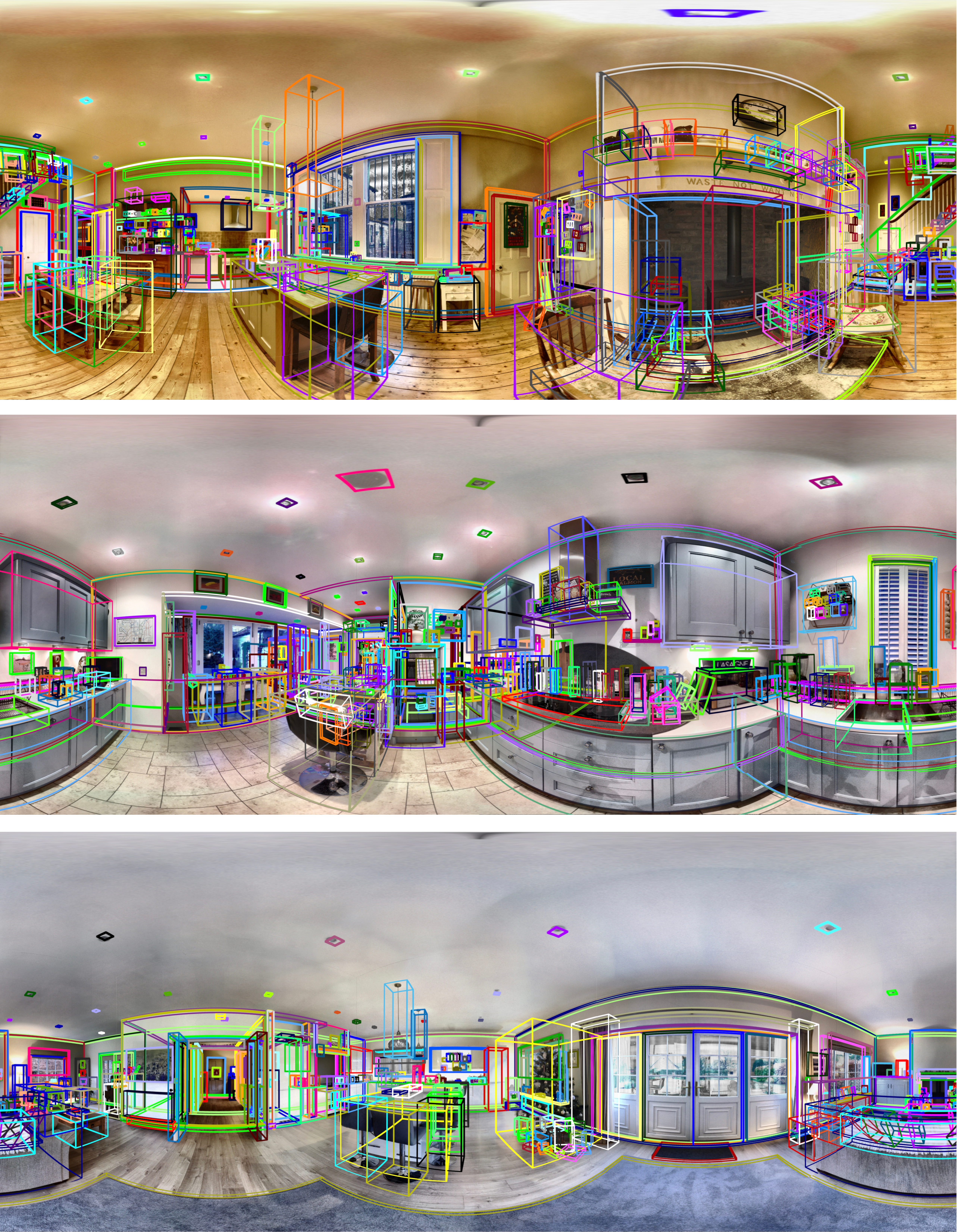}}
\label{fig:more_samples}
]

\end{document}